\documentclass[parallelisme]{compas2023}
\usepackage{color}
\usepackage{graphicx}                         
\usepackage{wrapfig}
\usepackage{booktabs}                         
\usepackage{multirow}
\usepackage{enumitem}                         
\usepackage{palatino}                         
\usepackage[font=small,labelfont=bf]{caption} 
\usepackage{multicol}                         
\usepackage{threeparttable,booktabs}          
\usepackage{soul}
\usepackage{xspace}
\usepackage{caption}                          
\usepackage{subcaption}                       
\definecolor{Paired-1}{RGB}{31,120,180}
\definecolor{Paired-2}{RGB}{166,206,227}
\definecolor{Paired-3}{RGB}{51,160,44}
\definecolor{Paired-4}{RGB}{178,223,138}
\definecolor{Paired-5}{RGB}{227,26,28}
\definecolor{Paired-6}{RGB}{251,154,153}
\definecolor{Paired-7}{RGB}{255,127,0}
\definecolor{Paired-8}{RGB}{253,191,111}
\definecolor{Paired-9}{RGB}{106,61,154}
\definecolor{Paired-10}{RGB}{202,178,214}
\definecolor{Paired-11}{RGB}{177,89,40}
\definecolor{Paired-12}{RGB}{255,255,153}

\newcommand{\meteorix}[0]{{\sc Meteorix}\xspace}
\newcommand{\source}[0]{{\sc Source}\xspace}

\newcommand{\rpi}[0]{Raspberry Pi 4\xspace}
\newcommand{\affect}[0]{{\sc AFF3CT}\xspace}

\toappear{1} 

\begin{document}

\title{Parallélisation d'une nouvelle application embarquée pour la détection
automatique de météores}
\shorttitle{FMDT}

\author{Mathuran Kandeepan, Clara Ciocan, Adrien Cassagne et Lionel Lacassagne} 

\address{Sorbonne Université, CNRS, LIP6, F-75005 Paris, France\\
prenom.nom@lip6.fr}

\date{\today}

\maketitle
\begin{sloppypar}
\begin{abstract}
  Cet article présente les techniques mises en \oe uvre pour paralléliser une
  nouvelle chaîne de traitement de l'image. L'application permet de détecter
  automatiquement les météores depuis un flux vidéo non stabilisé. La cible
  finale de cette chaîne de traitement est un système sur puce à faible
  consommation énergétique pouvant être embarqué sur un ballon-sonde, ou bien
  dans un avion. Les méthodes mises en place pour répondre aux contraintes de
  traitement temps réel ($\geq$ 25 images par seconde) tout en restant sur des
  niveaux de consommation énergétique faible ($\leq$ 10 Watts) sont présentées.
  En outre, la chaîne de traitement est découpée en graphe de tâches puis elle
  est parallélisée. Les résultats obtenus démontrent l'efficacité de la
  parallélisation. Par exemple, sur la \rpi, la chaîne de traitement atteint
  42~images par seconde et consomme moins de 6~Watts pour des séquences vidéos
  Full HD.
    

  \MotsCles{Vision par ordinateur, systèmes embarqués, parallélisme,
  multi-thread, pipeline.}
\end{abstract}

\section{Introduction}

Avec l'augmentation des missions spatiales, la détection de météores est
devenue nécessaire afin d'évaluer le flux d'objets rentrant dans l'atmosphère et
pour planifier en toute sécurité les activités spatiales.
Aujourd'hui, il est possible d'acquérir des heures de vidéos contenant des
météores en positionnant une caméra soit sur Terre et dirigée vers le ciel, soit
à bord d'un avion volant à haute
altitude~\cite{Vaubaillon2023_tauHerculeids_AA}, soit embarquée sur un
ballon-sonde~\cite{ocana2019balloon}. Cependant, les apparitions de météores
représentent une infime partie de la durée totale des enregistrements vidéos.
L'exploitation de ces vidéos ainsi que l'analyse image par image peut être
fastidieuse pour les astronomes. Une automatisation de cette détection est donc
nécessaire, d'où le développement de la chaîne de traitement d'images intitulée
Fast Meteor Detection Toolbox (FMDT)~\footnote{https://github.com/alsoc/fmdt}.

Ce projet universitaire, développé par Sorbonne Université en collaboration avec
l'Institut de Mécanique Céleste et de Calcul des  Éphémérides (IMCCE), a
plusieurs enjeux majeurs. Pour les astrophysiciens, étudier les météores peut
apporter des réponses concernant les propriétés des objets dans le système
solaire~\cite{space}. De plus, déployer un système embarqué à bas coût
détectant les météores dans l'atmosphère serait une avancée
significative.

En effet, envoyer un ballon-sonde équipé d'une caméra et d'un système sur puce
n'est pas très coûteux (quelques milliers d'euros). \textbf{La chaîne de
traitement doit être capable d'analyser un flux vidéo \emph{Full HD}
(1920$\times$1200 pixels) à 25 images/seconde} pour satisfaire la contrainte de
temps réel. Les ressources en énergie nécessaires au fonctionnement du système
sont aussi limitées. À titre d'exemple, dans un projet universitaire de
nanosatellite~\cite{Rambaux2019_ESA,Millet2022_Meteorix_COSPAR}, \textbf{la
puissance instantanée disponible pour la chaîne de traitement est inférieure à
10~Watts}. Pour répondre à ce défi, la chaîne de détection doit être rapide et
efficace.

Les architectures CPU modernes sont majoritairement multi-c\oe ur. \textbf{Il
est indispensable d'exploiter le parallélisme multi-thread} pour en tirer parti.
Il existe plusieurs modèles de programmation pour adresser ce type de
parallélisme. Le plus commun repose sur l'utilisation d'une bibliothèque de
threads (ex.: les threads {\sc Posix}). Il est aussi possible d'utiliser des
directives à la compilation avec {\sc OpenMP} par example. Cette solution a pour
avantage d'être très concise. Des langages comme {\sc OpenCL} permettent de
programmer une grille de threads et de s'abstraire de l'architecture (CPU, GPU,
FPGA, etc.). Enfin, il existe aussi des solutions spécifiques à  un domaine
(\emph{Domain Specific Langages} ou DSL en anglais). Les DSL proposent des
constructeurs parallèles spécifiques à une classe d'applications. Dans cet
article, la parallélisation repose sur de \textbf{la programmation à base de
tâches}, et plus précisément, c'est le DSL \affect~\cite{Cassagne2019a,
Cassagne2023} qui est utilisé. Ce dernier se présente sous la forme d'une
bibliothèque (DSL enfoui) et est conçu pour supporter les applications de type
\emph{streaming} (incluant le traitement vidéo). Enfin, \affect est une
bibliothèque C++ et il est naturel d'encapsuler FMDT qui est écrit en langage C.

\textbf{La contribution principale de cet article est la parallélisation
\emph{multi-thread} de la chaîne de traitement FMDT.} Dans un premier temps le
système est décrit en un graphe de tâches. Ensuite, plusieurs décompositions de
ce graphe de tâches sont étudiées. Enfin, deux techniques de parallélisation
sont implémentées et combinées :
\begin{enumerate}
  \item le travail à la chaîne (noté \emph{pipeline} par la suite),
  \item la réplication des tâches (aussi appelé parallélisme \emph{fork-join}
  dans la littérature anglaise).
\end{enumerate}
Différentes configurations d'affectations des threads aux c\oe urs physiques
sont évaluées.

Cet article est organisé comme suit. La Section~\ref{travaux} présente un état
de l'art des applications de détection de météores existantes. La
Section~\ref{contributions} introduit la version initiale de la chaîne de
détection, ainsi que les premières optimisations apportées pour satisfaire les
contraintes d'embarquabilité. La Section~\ref{res} décrit les expérimentations
effectuées et les résultats obtenus. Enfin, la Section~\ref{conclusion} conclut
sur les travaux présentés.

\section{Travaux connexes} \label{travaux}

Jusqu'à maintenant, la majorité des chaînes de traitement d’images développées
sont exécutées depuis le sol avec des caméras fixes~\cite{gural1997operational,
molau1999meteor,audureau2014freeture,colas2020fripon}. Un des problèmes de la
détection depuis la surface de la Terre est qu'elle est sensible aux
perturbations météorologiques. C'est pour cela qu'il est intéressant de déployer
des systèmes de détection depuis la haute atmosphère (au dessus des nuages).
Ainsi, la caméra est en mouvement, à l'inverse de la détection sur Terre.
Cela simplifie l'observation, mais complexifie la chaîne de traitement.

\meteorix~\cite{Rambaux2019_ESA} est le premier projet universitaire développé
par Sorbonne Université dédié à la détection de météores depuis l'espace à bord
d'un nanosatellite. Le projet est toujours en phase de définition du système.
Le nanosatellite doit embarquer une caméra dans le domaine du visible et un
ordinateur exécutant une chaîne de traitement d'images pour la détection en
temps réel. Avec une caméra dirigée vers la Terre, la chaîne de traitement
s'appuie des méthodes basées sur le flot optique. Cependant, la chaîne de
traitement implémentée n'est temps réel que jusqu'à la résolution HD (1280
$\times$ 720 pixels)~\cite{Millet2022_Meteorix_COSPAR} alors que la cible de cet
article est la résolution Full HD (1920 $\times$ 1080 pixels).

\source~\cite{liegibel2022meteor} est un autre projet de nanosatellite dédié à
la démonstration d'observations de météores depuis l'espace et à la
quantification du flux de météores. En utilisant également le flot optique, la
chaîne de traitement obtient de bons résultats de détection mais n'atteint pas
la cadence temps réel (1.25 images/seconde) sur des vidéos
HD~\cite{petri2022satellite}.

Dans les projets \meteorix et \source, la caméra est dirigée vers la Terre et
l'algorithme de flot optique permet de différencier plusieurs mouvements : la
rotation de la Terre, le mouvement en orbite de la caméra, les éclairs, les
villes et les objets entrant dans l'atmosphère (dont les météores et les débris
spatiaux). Si ce type d'algorithme est efficace, il est aussi coûteux en temps
de calcul. Dans cet article, contrairement à \meteorix et \source, la caméra
regarde au \emph{limbe} : son axe est tangentiel à la Terre, la scène observée
comprend l'espace ainsi que l'atmosphère où sont visibles les météores ainsi que
certaines étoiles.

\section{Contributions} \label{contributions}

\subsection{Description de la chaîne de traitement} \label{chaine}

\begin{figure}
  \vspace{-0.2cm}
  \begin{center}
  \begin{subfigure}{\textwidth}
  \centering
  \includegraphics[scale=0.33,keepaspectratio]{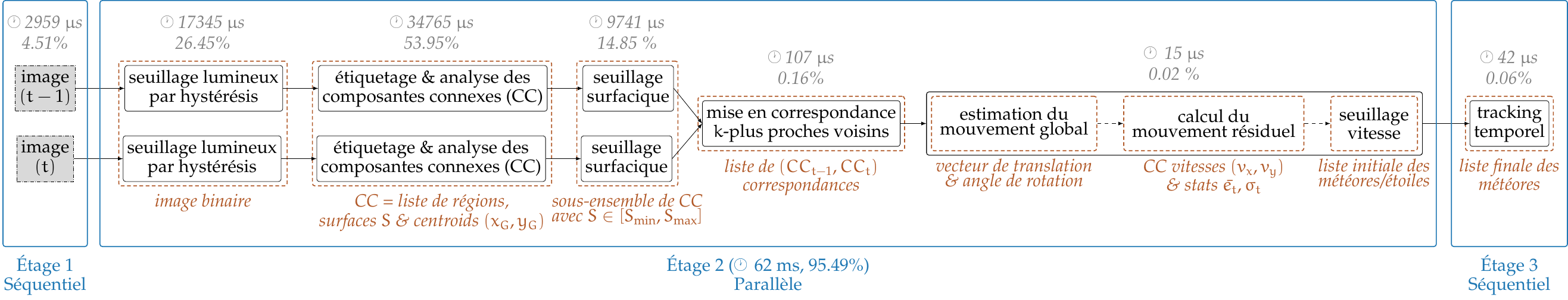}
  \subcaption{Chaîne de traitement initiale (version 1).}
  \label{fig:detection_chain_v1}
  \end{subfigure}
  \hfill
  \vspace{0.05cm}
  \begin{subfigure}{\textwidth}
  \centering
  \includegraphics[scale=0.33,keepaspectratio]{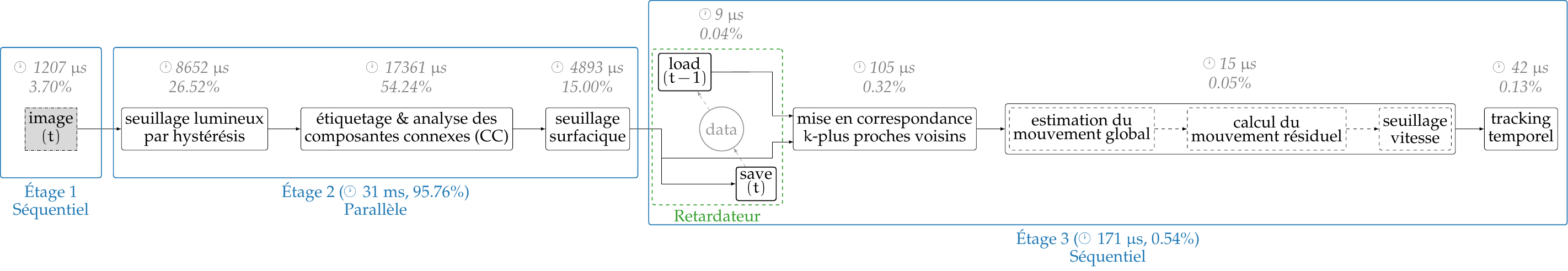}
  \subcaption{Chaîne de traitement optimisée (version 2).}
  \label{fig:detection_chain_v2}
  \end{subfigure}
  \caption{Différentes décompositions de la chaîne de détection de météores
  (avec les étages $E_1$, $E_2$ et $E_3$ du pipeline). La latence de chaque
  tâche est indiquée en gris (1 c\oe ur du \rpi en Full HD).}
  \label{fig:detection_chain}
  \end{center}
  \vspace{-1.0cm}
\end{figure}

La chaîne de traitement développée utilise différents algorithmes efficaces d'un
point de vue calculatoire. La Figure \ref{fig:detection_chain_v1} présente la
chaîne de traitement utilisée pour la détection de météores. Elle prend en
entrée un flux d'images en niveaux de gris et retourne en sortie la liste des
météores.

La première étape consiste à appliquer un seuillage lumineux sur toute l'image
pour distinguer les régions d'intérêt (météores, étoiles, planètes, satellites,
...). Un seuillage par hystérésis est utilisé. Ce dernier consiste en un
filtrage des pixels inférieurs à un seuil bas. Parmi les régions d'intérêt
restantes, seules celles contenant au moins un pixel supérieur à un seuil haut
sont retenues.

La deuxième étape consiste à passer d'une représentation binaire de l'image à
une liste de régions appelées des composantes connexes (CC) via un algorithme
d'étiquetage en composantes connexes (ECC). L'algorithme \textit{Light Speed
Labeling}~\cite{Lacassagne2009} est utilisé car il combine étiquetage et analyse
en composantes connexes (ACC). Dans notre cas, les caractéristiques des régions
sont la surface $S$ en pixels, le centre d'inertie de coordonnées $(x_G,y_G)$ et
le rectangle englobant ($[x_\text{min},x_\text{max}] \times
[y_\text{min},y_\text{max}]$) de chaque CC.

La troisième étape est un seuillage surfacique ne gardant que les CC dont la
surface est comprise entre $S_\text{min}$ et $S_\text{max}$. Ces trois premières
étapes sont appliquées sur deux images successives $I_{t-1}$ et $I_{t}$ pour
récupérer deux ensembles de CC.

La quatrième étape est la mise en correspondances (algorithme des
\textit{$k$-plus proches voisins} ou $k$-PPV) des CC de deux images
consécutives.
Pour chaque CC de l'image $I_t$ ayant trouvé une association dans $I_{t-1}$, un
vecteur distance est calculé.

La cinquième étape estime le mouvement global (un vecteur translation $\vec{T}$
et un angle de rotation $\theta$) de l'image du au balancement et au mouvement
de la caméra~\cite{bloch2020recalage}.

La sixième étape recale l'image $I_{t}$ sur l'image $I_{t-1}$. Après recalage,
les couples de CC dont la distances est quasi nulle représentent des étoiles. À
contrario, les couple de CC dont les distances sont significativement plus
grandes sont des régions en mouvement et potentiellement des météores. Ce
vecteur distance représente alors une estimation de la vitesse de l'objet.

La septième étape applique un seuillage en fonction de la vitesse de chaque CC
pour supprimer les objets immobiles (étoiles, planètes, etc.).

La huitième et dernière étape est le \textit{tracking} temporel. Si une CC est
détectée en mouvement sur au moins 3 images, alors une piste est créée.

\subsection{Optimisations}

\subsubsection{Décomposition en graphe de tâches}

Sur la Figure~\ref{fig:detection_chain_v1}, le seuillage lumineux, l'ECC, l'ACC
et le seuillage morphologique sont appliqués sur les images $I_{t-1}$ et
$I_{t}$. Or, à $t+1$, les mêmes traitements sont ré-appliqués sur l'image $I_t$.
On remarque donc que l'on calcule inutilement deux fois les CC sur l'image
$I_t$. Étant donné que ces traitements sont coûteux, il est préférable de ne pas
les recalculer et de mémoriser les CC pour l'itération suivante. La
Figure~\ref{fig:detection_chain_v2} illustre la nouvelle chaîne de traitement.
Un couple de tâches, appelé \emph{Retardateur}, a été ajouté entre le seuillage
morphologique et $k$-PPV. Après le seuillage, la tâche \emph{load} est exécutée.
Cette dernière lit les CC correspondants à l'image $I_{t-1}$. Ensuite, la tâche
\emph{save} est exécutée. Elle sauvegarde les CC correspondants à l'image
$I_{t}$ pour le traitement de la future image à $t+1$. Enfin, la tâche $k$-PPV
peux s'exécuter avec les bonnes entrées : les CC à $t-1$ et les CC à $t$.

Dans la suite de cet article, le graphe de tâches correspondant à la
Figure~\ref{fig:detection_chain_v1} est référencé comme la version 1 et le
graphe de tâches correspondant à la Figure~\ref{fig:detection_chain_v2} est
référencé comme la version 2. Sur les deux graphes de tâches, le pourcentage de
temps et le temps de chaque tâche est indiqué. Ces temps ont été mesurés depuis
une exécution séquentielle sur la \rpi. Pour la version 1, la latence est de
65~ms alors que pour la version 2, la latence est de 32~ms. La version 2 est
donc 2$\times$ plus rapide que la version 1.

\subsubsection{Parallélisation du graphe de tâches}

\paragraph{Pipeline}

\begin{figure}
  \vspace{-0.2cm}
  \begin{center}
  \begin{subfigure}{\textwidth}
  \centering
  \includegraphics[width=0.95\textwidth]{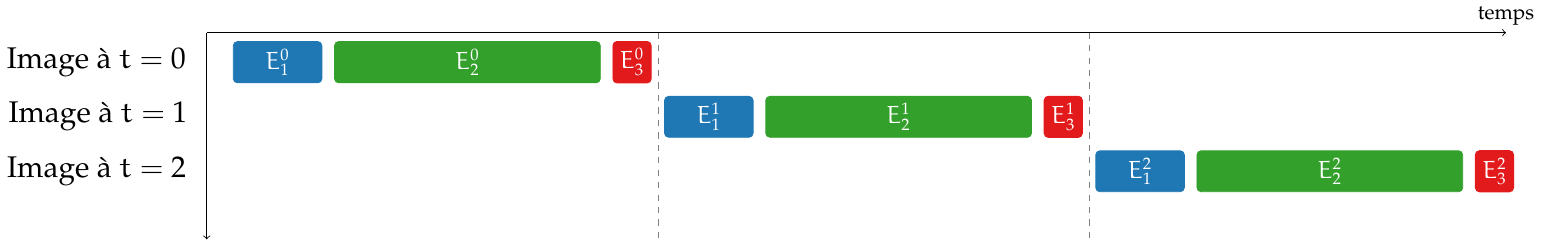}
  \subcaption{Exécution séquentielle.}
  \label{fig:seq}
  \end{subfigure}
  \hfill
  \begin{subfigure}{\textwidth}
  \centering
  \includegraphics[width=0.95\textwidth,keepaspectratio]{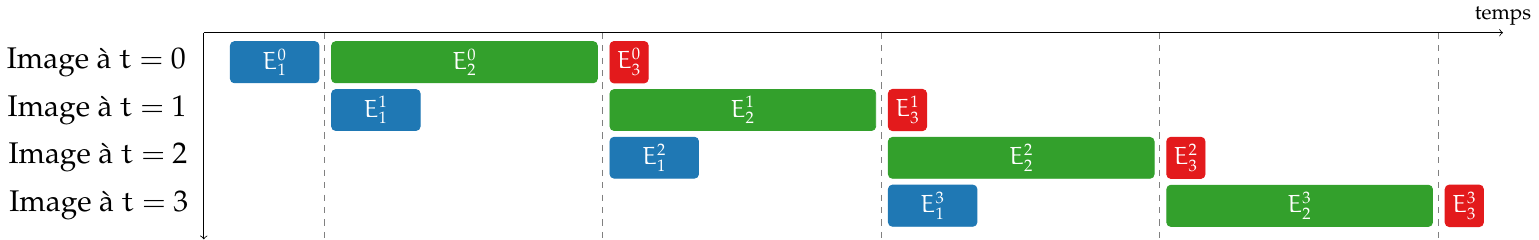}
  \subcaption{Exécution à la chaîne (parallélisme \emph{pipeline}).}
  \label{fig:pip2}
  \end{subfigure}
  \caption{Exécutions séquentielle et pipelinée (avec $E_1^t$, $E_2^t$ et
  $E_3^t$, les étages 1, 2 et 3 à l'instant $t$).}
  \label{fig:seq_pip}
  \end{center}
  \vspace{-1.0cm}
\end{figure}

En séquentiel, les tâches de la chaîne de traitement sont exécutées les unes
après les autres. Ce type d'exécution est illustré par la Figure~\ref{fig:seq}.
Dans un contexte multi-c\oe ur, il est possible de regrouper des tâches
consécutives pour former des \emph{étages} et appliquer le principe du travail à
la chaîne (\emph{pipeline}). Lors d'une exécution pipelinée, l'étage
$E_{e}^{t+1}$ peut s'exécuter en même temps que l'étage $E_{e+1}^{t}$ (avec $e$
le numéro de l'étage et $t$ le numéro temporel de l'image à traiter). Ce type de
parallélisme a pour avantage de conserver les dépendances de données. En régime
permanent, cela se traduit par une exécution en parallèle de tous les étages sur
différents threads (voir Figure~\ref{fig:pip2}). Le débit théorique est alors
celui de l'étage le plus lent. La Figure~\ref{fig:seq_pip} montre que
l'exécution pipelinée atteint un débit plus élevé que  l'exécution séquentielle.
Dans ce travail, la chaîne de traitement a été découpée en 3 étages (voir
Figure~\ref{fig:detection_chain}).

\paragraph{Réplication de tâches}\label{sec:dup}

Dans la chaîne de détection étudiée, certaines tâches peuvent être exécutées en
parallèle sur différents threads et sur différentes images. C'est notamment le
cas des tâches de l'étage $E_2$ dans la Figure~\ref{fig:detection_chain}. C'est
ce que l'on appelle le parallélisme de données.

Ainsi, il est possible d'exécuter des tâches sur plusieurs c\oe urs, de sorte
que l'exécution se ramifie à certain points du graphe pour se rejoindre et
reprendre l'exécution séquentielle ensuite. C'est ce qu'on appelle la
réplication de tâches. En ramifiant l'exécution de l'étage $E_2$, soit l'étage
dont la latence est la plus élevée, il est possible d'augmenter le débit.

\paragraph{Implémentation avec la bibliothèque AFF3CT}

\affect permet de décrire des applications selon le formalisme de \emph{flux de
données}. Pour cela, plusieurs composants sont définis : la \textit{séquence},
la \textit{tâche} et la \textit{socket}. Une tâche est un traitement effectué
sur des données. Chaque étape de la chaîne de traitement est une tâche. Les
tâches sont caractérisées par des sockets. Une socket définit un chemin par
lequel des tâches peuvent consommer et/ou produire des données. Une séquence est
un ensemble de tâches exécutées avec un ordonnancement déterminé à sa
construction.

De plus, \affect vient avec un support d'exécution multi-thread comprenant une
implémentation du pipeline pouvant être combinée à de la réplication automatique
de séquence. Cela à la condition que toutes les tâches qui composent la séquence
soient sans état interne. Une séquence correspond à un étage du pipeline.

Dans ces travaux, un pipeline à 3 étages a été instancié. Et, dans le second
étage du pipeline, la réplication de tâches a été activée. Les synchronisations
entre l'étage~1-2 et entre l'étage~2-3 ont été configurées avec de l'attente
passive, permettant ainsi de maximiser l'utilisation des ressources au détriment
d'un léger surcoût en latence. Lorsqu'un thread s'endort, le système
d'exploitation peut ré-affecter le c\oe ur correspondant à un autre thread
actif. Enfin, la synchronisation entre les étages du pipeline est assurée par un
algorithme de type producteur-consommateur. Il y a des \emph{buffers} pour
l'échange de données entre les threads. La taille de ces \emph{buffers} a été
fixée à 1, permettant ainsi de minimiser la latence sans qu'il n'y ait d'impact
significatif sur le débit. Cela est vrai pour un petit nombre de c\oe urs
homogènes ($n \leq 4$).

\section{Expérimentations et résultats} \label{res}

\begin{table}[t]
  \vspace{-0.2cm}
  \center
  {\resizebox{0.95\linewidth}{!}{
  \begin{tabular}{r r r r r r r r }
  \toprule
                  Réf. &                            Nom complet  &                 Date  &               Gravure  &                                         CPUs &             Fréq. &            \multicolumn{2}{r}{RAM (taille \& débit)} \\
  \midrule
  \multirow{2}{*}{XU4} &  \multirow{2}{*}{Hardkernel Odroid-XU4} & \multirow{2}{*}{2016} & \multirow{2}{*}{28 nm} &  4 $\times$ \textit{LITTLE} ARMv7 Cortex-A7  &           1.4~GHz & \multirow{2}{*}{ 2 GB} & \multirow{2}{*}{  3.5 GB/s} \\
                       &                                         &                       &                        &  4 $\times$    \textit{Big} ARMv7 Cortex-A15 &           1.5~GHz &                        &                             \\ \addlinespace
                 RPi4  &                           \rpi model B  &                 2019  &                 28 nm  &  4 $\times$    \textit{Big} ARMv8 Cortex-A72 &           1.5 GHz &                  8 GB  &                   3.9 GB/s  \\ \addlinespace
                 Nano  &                     Nvidia Jetson Nano  &                 2019  &                 20 nm  &  4 $\times$    \textit{Big} ARMv8 Cortex-A57 & $\approx$ 1.5 GHz &                  4 GB  &                   9.0 GB/s  \\ \addlinespace
  \multirow{2}{*}{ M1} & \multirow{2}{*}{Apple Silicon M1 Ultra} & \multirow{2}{*}{2022} &  \multirow{2}{*}{5 nm} &  4 $\times$ \textit{E-core} ARMv8 Icestorm   & $\approx$ 2.0~GHz & \multirow{2}{*}{64 GB} & \multirow{2}{*}{344.0 GB/s} \\
                       &                                         &                       &                        & 16 $\times$ \textit{P-core} ARMv8 Firestorm  & $\approx$ 3.0~GHz &                        &                             \\
  \bottomrule
  \end{tabular}
  }}
  \caption{Spécifications des différents SoC évalués.}
  \label{tab::archi}
  \vspace{-0.5cm}
\end{table}

La Table~\ref{tab::archi} présente les différentes architectures embarquées
évaluées. La XU4 et le M1 sont hétérogènes et disposent : (1) de c\oe urs
efficaces du point de vue énergétique (\emph{LITTLE} ou \emph{E-core}) et (2) de
c\oe urs puissants/rapides (\emph{Big} ou \emph{P-core}). Le code est compilé
avec GCC v9.4.0 et avec les options d'optimisations suivantes :
\texttt{-O3 -funroll-loops -march=native}. Toutes les expérimentations ont été
faites sur une vidéo Full HD où 100\% des météores sont
détectés~\cite{Vaubaillon2023_tauHerculeids_AA}. Enfin, seule la version 2 de la
chaîne de traitement est considérée (voir Figure~\ref{fig:detection_chain_v2}).

Les résultats obtenus sont étudiés au travers de 4 métriques :
le {\color{Paired-1} débit moyen} $\mathcal{D}$ en nombre d'images par seconde,
la {\color{Paired-5}latence moyenne} $\mathcal{L}$ pour traiter une image,
la {\color{Paired-3}puissance moyenne} $\mathcal{P}$ en Watts,
la {\color{Paired-7}consommation énergétique moyenne} $\mathcal{E}$ pour traiter
une image en milliJoules (mJ).

\begin{figure}[t]
    \centering
    \includegraphics[width=1\textwidth]{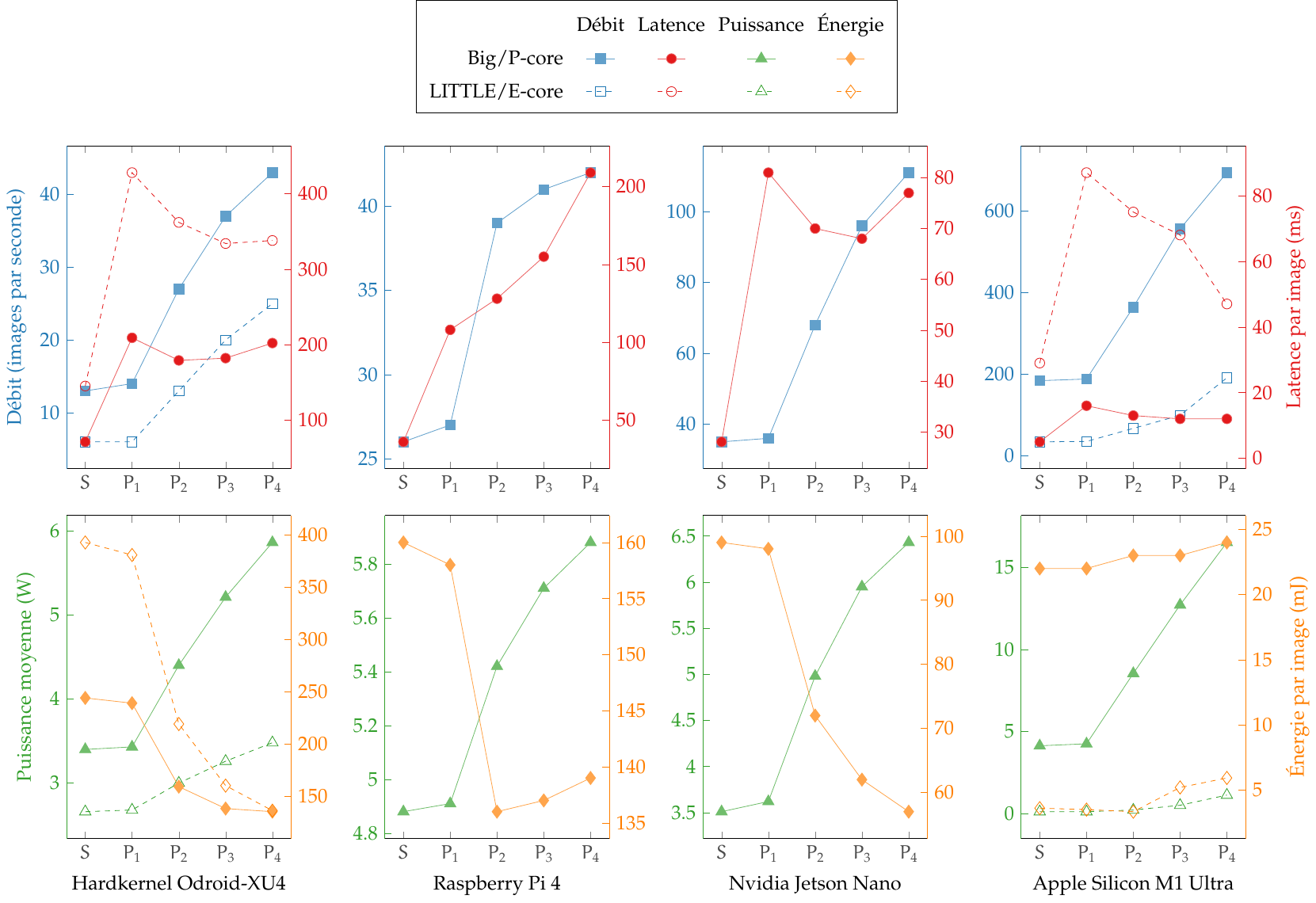}
    \caption{Performances Full HD de la chaîne de détection (version 2, c.f.
    Figure~\ref{fig:detection_chain_v2}) en terme de débit, latence, puissance
    moyenne et énergie consommée pour quatre cibles embarquées.}
    \label{fig:results}
    \vspace{-0.8cm}
\end{figure}

La Figure~\ref{fig:results} montre les résultats obtenus selon les 4 métriques
énoncées précédemment. La complexité calculatoire ne varie pas (ou très peu) en
fonction des images. En effet, les scènes observées sont presque identiques :
quelques petites régions lumineuses (les étoiles ou les météores) sur un fond
majoritairement foncé. Pour que les mesures soient significatives, chaque
expérience est exécutée pendant 30~secondes.

$\mathcal{D}$, $\mathcal{L}$, $\mathcal{P}$ et
$\mathcal{E}$ sont étudiées en fonction de l'exécution séquentielle (notée $S$)
et de plusieurs exécutions pipelinées (notées $P_i$) de la chaîne de traitement.
Lors d'une exécution pipelinée, 1 thread est affecté à l'étage $E_1$ et un autre
thread est affecté à l'étage $E_3$. Le nombre de threads de l'étage $E_2$ peut
varier grâce à la réplication (voir Section~\ref{sec:dup}). Ainsi, $i$ indique
le nombre de threads affectés à l'étage $E_2$. Par exemple $P_3$ signifie qu'il
y a 1 thread pour l'étage $E_1$, 3 threads pour l'étage $E_2$ et 1 thread pour
l'étage $E_3$ (soit 5 threads au total). En fonction du nombre de c\oe urs
physiques de la cible, il peut y avoir plus de threads que de c\oe urs
physiques.

\paragraph{Débit et latence}

La Figure~\ref{fig:results} montre que le pipeline $P_1$ n'apporte pas un gain
significatif en terme de débit par rapport à la version séquentielle $S$. De
plus, la latence de $P_1$ est triplée par rapport à une exécution séquentielle.
Pour la détection de météores, la latence n'est pas critique tant que l'on
peut mettre quelques images ($< 10$) en attente dans une mémoire tampon (la
taille de la RAM des SoC évalués est largement suffisante). Pour des instances
de $P_i$ avec $i \geq 2$, les débits sont nettement meilleurs que la version
$S$. La réplication dans l'étage $E_2$ du pipeline a un impact direct sur le
débit final. En fonction des cibles, la latence peut soit diminuer légèrement
lorsque $i$ grandit (XU4, Nano et M1) soit, au contraire, augmenter (RPi4). Sur
la RPi4, la bande passante de la mémoire globale est trop faible pour alimenter
efficacement plus de 2 c\oe urs dans l'étage $E_2$. Pour toutes les cibles
étudiées, il existe au moins une configuration qui tient la cadence temps réel
de 25 images par seconde.

\paragraph{Puissance et consommation énergétique}

Les mesures de puissance $\mathcal{P}$ sont faites à la prise d'alimentation
pour la XU4, la RPi4 et la Nano. Pour le M1, c'est l'outil \texttt{powermetrics}
d'Apple qui est utilisé. Ce dernier mesure uniquement la consommation du CPU.
L'énergie $\mathcal{E}$ est fonction de la puissance moyenne $\mathcal{P}$ et du
débit $\mathcal{D}$. Sur la Figure~\ref{fig:results}, on constate que
généralement quand le nombre de c\oe urs utilisé augmente alors la puissance
moyenne $\mathcal{P}$ augmente. C'est ce qui est attendu. Pour la consommation
énergétique, on observe plutôt la tendance inverse : plus il y a de ressources,
moins on consomme d'énergie. Cependant, cela n'est pas toujours vrai, par
exemple pour la RPi4 et pour le M1, on perd en efficacité à partir de 3 c\oe urs
dans l'étage $E_2$. Comme la contrainte en débit est respectée pour 2 c\oe urs,
il n'est pas intéressant d'affecter plus de c\oe urs à cet étage. À l'exception
des \emph{P-cores} du M1, toutes les cibles respectent la contrainte de
puissance fixée à 10 Watts.

\section{Conclusion et perspectives} \label{conclusion}

Dans cet article, des méthodes de parallélisation (pipeline et réplication) ont
été présentées puis implémentées sur une nouvelle chaîne de détection de
météores. L'application a été évaluée sur plusieurs cibles embarquées. Grâce à
l'implémentation multi-threads, toutes les cibles respectent les contraintes en
terme de débit et de puissance sur des séquences vidéos Full HD.

Sur certaines cibles (Nvidia Jetson Nano et Apple M1 Ultra), il est
envisageable de traiter des résolutions plus importantes comme la QHD (2560
$\times$ 1440 pixels). Dans des travaux futurs, il serait possible de combiner
du code CPU SIMD~\cite{Lemaitre2020_SIMD_CCL_WPMVP} pour l'étiquetage en
composantes connexes avec de la programmation sur GPU pour certains traitements
réguliers comme les seuillages.

\newpage
\bibliography{refs}

\def\No{\kern-.25em\lower.2ex\hbox{\char'27}}
\begin{thebibliography}{10}

\bibitem{audureau2014freeture}
Audureau (Y.), Marmo (C.), Bouley (S.), Kwon (M.-K.), Colas (F.), Vaubaillon
  (J.), Birlan (M.), Zanda (B.), Vernazza (P.), Caminade (S.) et~al. --
\newblock Freeture: A free software to capture meteors for fripon. --
\newblock In {\em International Meteor Conference}, pp. 39--41, 2014.

\bibitem{bloch2020recalage}
Bloch (I.). --
\newblock Recalage et fusion d’images m{\'e}dicales, 2020.
  https://perso.telecom-paristech.fr/bloch/P6Image/recalageImMed.pdf.

\bibitem{space}
C.~Hongru (N.~R.) et Vaubaillon (J.). --
\newblock Accuracy of meteor positioning from space- and ground-based
  observations. 2020.

\bibitem{Cassagne2019a}
Cassagne (A.), Hartmann (O.), L\'eonardon (M.), He (K.), Leroux (C.), Tajan
  (R.), Aumage (O.), Barthou (D.), Tonnellier (T.), Pignoly (V.), {Le Gal} (B.)
  et J\'ego (C.). --
\newblock {AFF3CT}: A fast forward error correction toolbox! {\em Elsevier
  SoftwareX}, vol.~10, 2019, p. 100345.

\bibitem{Cassagne2023}
Cassagne (A.), Tajan (R.), Aumage (O.), Barthou (D.), Leroux (C.) et J\'ego
  (C.). --
\newblock A {DSEL} for high throughput and low latency software-defined radio
  on multicore {CPU}s. {\em Wiley Concurrency and Computation: Practice and
  Experience (CCPE)}, juillet 2023, p. e7820.

\bibitem{colas2020fripon}
Colas (F.), Zanda (B.), Bouley (S.), Jeanne (S.), Malgoyre (A.), Birlan (M.),
  Blanpain (C.), Gattacceca (J.), Jorda (L.), Lecubin (J.) et~al. --
\newblock Fripon: a worldwide network to track incoming meteoroids. {\em
  Astronomy \& Astrophysics}, vol.~644, 2020, p. A53.

\bibitem{gural1997operational}
Gural (P.~S.). --
\newblock An operational autonomous meteor detector: Development issues and
  early results. {\em WGN, Journal of the International Meteor Organization},
  vol.~25, 1997, pp. 136--140.

\bibitem{Lacassagne2009}
Lacassagne (L.) et Zavidovique (B.). --
\newblock {L}ight {S}peed {L}abeling for {RISC} architectures. --
\newblock In {\em International Conference on Image Analysis and Processing},
  pp. 3245--3248. IEEE, 2009.

\bibitem{Lemaitre2020_SIMD_CCL_WPMVP}
Lemaitre (F.), Hennequin (A.) et Lacassagne (L.). --
\newblock How to speed {C}onnected {C}omponent {L}abeling up with {SIMD} {RLE}
  algorithms. --
\newblock In {\em Workshop on Programming Models for {SIMD}/Vector Processing},
  pp. 1--8. ACM, 2020.

\bibitem{liegibel2022meteor}
Liegibel (M.), Petri (J.), Hoffmann (P.), Geier (N.) et Klinkner (S.). --
\newblock Meteor observation with the source cubesat--developing a simulation
  to test on-board meteor detection algorithms. --
\newblock In {\em Symposium on Space Educational Activities}. Universitat
  Polit{\`e}cnica de Catalunya, 2022.

\bibitem{Millet2022_Meteorix_COSPAR}
Millet (M.), Rambaux (N.), Cassagne (A.), Bouyer (M.), Petreto (A.) et
  Lacassagne (L.). --
\newblock High performance computer vision application for {M}eteor detection
  from a cubesat. --
\newblock In {\em Committee on Space Research}, 2022.

\bibitem{molau1999meteor}
Molau (S.). --
\newblock The meteor detection software metrec. --
\newblock In {\em International Meteor Conference}, pp. 9--16, 1999.

\bibitem{ocana2019balloon}
Oca{\~n}a (F.), de~Miguel (A.~S.), Project (D.) et~al. --
\newblock Balloon-borne video observations of geminids 2016, 2019.
  arXiv:1911.10064.

\bibitem{petri2022satellite}
Petri (J.). --
\newblock {\em Satellite Formation and Instrument Design for Autonomous Meteor
  Detection}. --
\newblock phdthesis, Universit{\"a}t Stuttgart, 2022.

\bibitem{Rambaux2019_ESA}
Rambaux (N.), Vaubaillon (J.), Lacassagne (L.), Galayko (D.), Guignan (G.),
  Birlan (M.), Capderou (M.), Colas (F.), Deleflie (F.), Deshours (F.),
  Hauchecorne (A.), Keckhut (P.), Levasseurd-Regourd (A.), Rault (J.) et Zanda
  (B.). --
\newblock Meteorix: a cubesat mission dedicated to the detection of meteors and
  space debris. --
\newblock In {\em ESA Space Safety Programme Office, NEO and Debris Detection
  Conference}, pp. 1--9, 2019.

\bibitem{Vaubaillon2023_tauHerculeids_AA}
Vaubaillon (J.), Loir (C.), Ciocan (C.), Kandeepan (M.), Millet (M.), Cassagne
  (A.), Lacassagne (L.), da~Fonseca (P.), Zander (F.), Buttsworth (D.), Loehle
  (S.), T{\'o}th (J.), Gray (S.), Moingeon (A.) et Rambaux (N.). --
\newblock A 2022 $\tau$-herculid meteor cluster from an airborne experiment:
  Automated detection, characterization, and consequences for meteoroids. {\em
  Astronomy and Astrophysics}, 2023.

\end{thebibliography}
\end{sloppypar}
\end{document}